%% file: main.tex
\pdfoutput=1

\documentclass[11pt]{article}

\usepackage[]{EACL2023}

\usepackage{times}
\usepackage{latexsym}

\usepackage[T1]{fontenc}

\usepackage[utf8]{inputenc}

\usepackage{microtype}
\usepackage{adjustbox}

\newif\ifcomments
\commentstrue
\ifcomments
    
    \providecommand\dr[1]{\textcolor{blue}{[DR: #1]}}
    \providecommand\rd[1]{\textcolor{purple}{[RD: #1]}}
    \providecommand\todo[1]{\textcolor{red}{[TODO: #1]}}
\else
    \providecommand{\hw}[1]{}
    \providecommand{\dr}[1]{}
    \providecommand{\rd}[1]{}
    \providecommand{\todo}[1]{}
\fi

\usepackage{graphicx}
\usepackage{caption}
\usepackage{subcaption}
\usepackage{amsmath}
\usepackage{amsthm}
\usepackage{amssymb}
\usepackage{resizegather}
\usepackage{multicol}
\usepackage{soul}
\usepackage{xcolor}
\usepackage{xspace}
\usepackage{mathtools}
\usepackage{bm}
\usepackage{multirow}
\usepackage{booktabs}
\usepackage{esvect}
\usepackage{amsfonts}
\usepackage{cleveref}
\usepackage{footmisc}
\usepackage{smartdiagram}
\usesmartdiagramlibrary{additions} 
\usetikzlibrary{arrows,quotes,calc,decorations.text,positioning,fit} 
\usepackage{tikz}
\usepackage{geometry}
\newcommand{\hlc}[2][yellow]{ {\sethlcolor{#1} \hl{#2}} }
\newcommand{\stitle}[1]{\vspace{1ex}\noindent{\bf #1}}
\makeatletter
\newcommand{\printfnsymbol}[1]{%
  \textsuperscript{\@fnsymbol{#1}}%
}
\makeatother
\crefformat{section}{\S#2#1#3}
\crefformat{subsection}{\S#2#1#3}
\crefformat{subsubsection}{\S#2#1#3}
\crefrangeformat{section}{\S\S#3#1#4 to~#5#2#6}
\crefmultiformat{section}{\S\S#2#1#3}{ and~#2#1#3}{, #2#1#3}{ and~#2#1#3}
\usepackage{refstyle}
\Crefformat{figure}{#2Fig.~#1#3}
\Crefmultiformat{figure}{Figs.~#2#1#3}{ and~#2#1#3}{, #2#1#3}{ and~#2#1#3}
\Crefformat{table}{#2Tab.~#1#3}
\Crefmultiformat{table}{Tabs.~#2#1#3}{ and~#2#1#3}{, #2#1#3}{ and~#2#1#3}
\Crefformat{appendix}{\S#2#1#3}
\crefformat{algorithm}{Alg.~#2#1#3}
\Crefformat{equation}{Eq.~#2#1#3}

\title{ \vspace*{-0.5in}
{{\small \hfill EACL'23 Findings}\\
\vspace*{.25in}} Zero-Shot On-the-Fly Event Schema Induction}

\author{Rotem Dror\thanks{\;\;Indicating equal contribution.}\;, Haoyu Wang$^{*}$, and Dan Roth \\
        Department of Computer and Information Science \\ 
        University of Pennsylvania \\
        \texttt{\{rtmdrr,why16gzl,danroth\}}@seas.upenn.edu}

\begin{document}
\maketitle

\input{00_abstract}
\input{01_introduction}
\input{02_related_work}
\input{03_data_generation}

\input{04_information_extraction}
\input{05_schema_induction}
\input{06_experiments}

\input{09_conclusions}
\input{08_ack}
\input{11_limitations}
\input{10_ethics}

\bibliography{custom, anthology}
\bibliographystyle{acl_natbib}

\clearpage
\appendix
\input{appendix_turing_experiment}

\input{App1_turing_results}
\input{app2_instructions}
\input{wizard_of_oz}

\end{document}

%% file: 00_abstract.tex
\begin{abstract}
What are the events involved in a pandemic outbreak? What steps should be taken when planning a wedding? The answers to these questions can be found by collecting many documents on the complex event of interest, extracting relevant information, and analyzing it. We present a new approach\footnote{\url{https://cogcomp.seas.upenn.edu/page/publication_view/995}} 
in which large language models are utilized to generate source documents that allow predicting, given a high-level event definition, the specific events, arguments, and relations between them to construct a schema that describes the complex event in its entirety.
Using our model, complete schemas on any topic can be generated on-the-fly without any manual data collection, i.e., in a zero-shot manner. Moreover, we develop efficient methods to extract pertinent information from texts and demonstrate in a series of experiments that these schemas are considered to be more complete than human-curated ones in the majority of examined scenarios. Finally, we show that this framework is comparable in performance with previous supervised schema induction methods that rely on collecting real texts and even reaching the best score in the prediction task.
\end{abstract}

%% file: 01_introduction.tex
\section{Introduction}
\label{01_introduction}

Event processing refers to tracking, analyzing, and drawing conclusions from streams of information about events. This event analysis aims at identifying meaningful events (such as opportunities or threats) in real-time situations and responding appropriately. Event processing can also be utilized to gain a deep understanding of the specific steps, arguments, and relations between them that are involved in a complex event. The information above can be consolidated into a graphical representation called an \textit{event schema} \cite{li-etal-2021-future}.
For instance in \Cref{fig:kidnapping}, the graph representation of events and participants assists in gaining an understanding 
of the complex event of kidnapping and could help composing 
a reaction plan if needed. 
\input{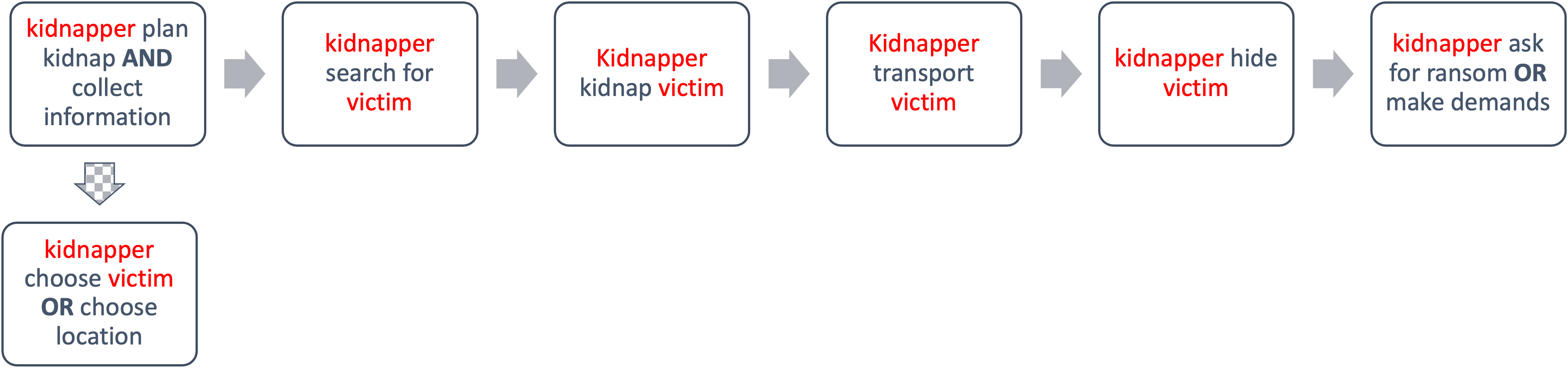}

The NLP community has devoted much effort to understanding events that are described in a document or in a collection of documents for this purpose. These efforts include identifying event triggers \cite{lu-roth-2012-automatic, huang-etal-2018-zero, wadden-etal-2019-entity, han-etal-2019-joint}, extracting event arguments \cite{punyakanok-etal-2008-importance, peng-etal-2016-event, lin-etal-2020-joint, zhang-etal-2021-zero}, and predicting the relations between events, e.g., temporal, coreferential, causal or hierarchical relations \cite{do-etal-2012-joint, lee-etal-2012-joint, glavas-etal-2014-hieve, ning-etal-2018-joint, wang-etal-2020-joint, zhang-etal-2020-analogous, trong2022selecting}. 


Previous works on event schema induction relied on the information extracted from manually collected documents to build the schema graph. For instance, \citet{li-etal-2020-connecting} learn an auto-regressive language model (LM) over paths in the instance graphs depicting events, arguments and relations of instances of the complex events, and then construct a schema graph by merging the top $k$ ranked paths. Their approach, however, requires access to many documents on each topic of interest,
which can be extremely laborious and time consuming to obtain.

In this paper, our goal is to allow creating schemas on-the-fly by taking as input only the name of the complex event of interest (like a ``pandemic outbreak'' or an ``armed robbery''). 
To avoid manually collecting many documents on the topic of the schema, we utilize pre-trained text generators, e.g., GPT-3 \cite{brown2020language}, to obtain documents of diverse genres on the desired topic (examples presented in \Cref{fig:all_generated}). 
These documents are then processed to extract pertinent information from which a schema is constructed.
The fact that we do not collect any data makes our learning framework zero-shot since we do not rely on any human-collected articles or example schemas. 

In addition to eliminating the need to collect data, we also made the information extraction process faster by implementing new and efficient methods for identifying temporal and hierarchical relations between events mentioned in the text. 
These two steps are the most time consuming in the process of schema induction and could take up to 2 hours each using state-of-the-art models proposed by \citet{zhou-etal-2021-temporal, wang-etal-2021-learning-constraints}. 
Sending the whole text as input instead of two sentences at each time, our proposed model shortens the inference time significantly to several minutes without enduring a major loss in performance. 


The process of generating texts is explained in Section~\Cref{03_data_generation}, and the process of extracting relevant and salient information is described in Section~\Cref{04_information_extraction}, then we introduce the construction of schema graphs in Section~\Cref{05_schema_induction}. To evaluate our zero-shot schema generator we conduct experiments on a benchmark dataset for schema induction, LDC2020E25, and provide a new dataset for further evaluation called Schema-11. Additionally, we design a subject-matter expert Turing test, a.k.a. Feigenbaum test \cite{feigenbaum2003some}, to determine whether our algorithm could mimic experts' response. We also demonstrate that documents generated by GPT-3 are informative and useful for the task of schema induction. The experiments and results are presented in Section~\Cref{06_experiments}.
The contributions of our work include:
\begin{enumerate}
    \item Predicting an entire schema given the name of a complex event without collecting data.
    \item Implementing a novel and efficient One-Pass approach for identifying temporal and hierarchical relations between events.
    \item Presenting a method for automatically inducing logical relations between events based on temporal relations.
    \item Offering a Feigenbaum test for evaluation on a new schema dataset, Schema-11.
\end{enumerate}



%% file: figures/kidnapping.tex
\begin{figure*}
    \centering
    \scalebox{0.8}{
    \begin{tikzpicture}[]  
\tikzset{
    mynode/.style={rectangle,rounded corners, draw=gray, top color=white, bottom color=gray!40,very thick, inner sep=1em,
                  minimum size=1em, text centered,
                  drop shadow, text width=1.75cm},
    mytemp/.style={-stealth,color=gray!40, draw, line width=2pt},
    myhier/.style={-stealth,color=gray, draw, dashed, line width=2pt},
}
\node[mynode] (0) {Preparation};
\node[mynode, right=7cm of 0] (1) {Kidnapping};
\node[mynode, below left=1cm and -1cm of 0] (2) {Kidnapper plans \textbf{AND} collects \\ information};
\node[mynode, right=0.5cm of 2] (3) {Kidnapper looks for victim};
\node[mynode, below=0.5cm of 2](4){Kidnapper chooses victim \textbf{OR} location};
\node[rectangle,rounded corners,fit={(2) (3) (4)}, draw=gray, very thick] (outer1) {};
\node[mynode, below left=1.5cm and 2cm of 1](5){Kidnapper kidnaps victim};
\node[mynode, right=0.5cm of 5](6){Kidnapper transports victim};
\node[mynode, right=0.5cm of 6](7){Kidnapper hides victim};
\node[mynode, right=0.5cm of 7](8){Kidnapper asks for ransom \textbf{OR} makes demands};
\node[rectangle,rounded corners,fit={(5) (6) (7) (8)}, draw=gray, very thick] (outer2) {};
\path[mytemp] (0) to (1);
\path[mytemp] (2) to (3);
\path[myhier] (2) to (4);
\path[mytemp] (5) to (6);
\path[mytemp] (6) to (7);
\path[mytemp] (7) to (8);
\path[myhier] (0) to (outer1);
\path[myhier] (1) to (outer2);
\end{tikzpicture} 
}
    \caption{An example schema for the event of Kidnapping. The regular arrows represent temporal relations and the dashed arrows represent hierarchical relations (\textsc{Parent-Child}).}
    \label{fig:kidnapping}
\end{figure*}
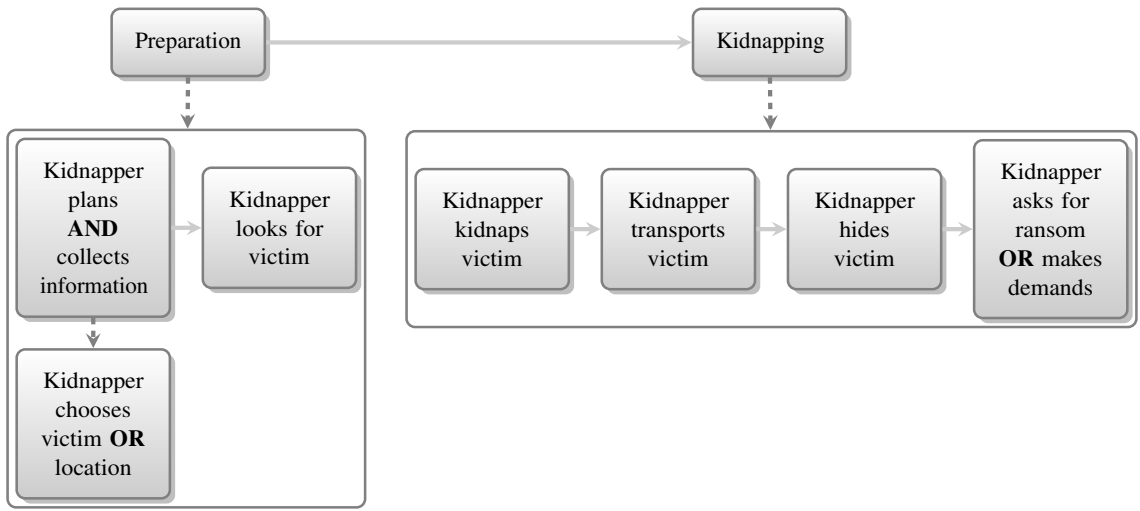

%% file: 02_related_work.tex
\section{Related Work}
\label{01_related_work}
\paragraph{Schema Induction:} Early schema induction efforts focused on identifying the triggers and participants of atomic events without considering relations between atomic events that comprise complex schemas \cite{chambers-2013-event, cheung-etal-2013-probabilistic, nguyen-etal-2015-generative, sha-etal-2016-joint, yuan2018open}. More recent work focuses on inducing schemas for pairs of events \cite{li-etal-2020-connecting} and multiple events \cite{zhang-etal-2021-learning, li-etal-2021-future}, but they require access to large corpora for the induction process. 
In this work, we induce schemas on-the-fly in a zero-shot manner. 
As is standard in state-of-the-art (SOTA) works \cite{li-etal-2020-connecting,li-etal-2021-future, wen-etal-2021-resin, lawley2022mining}, we output all the essential information about relations between events and arguments extracted from the text, in addition to logical and hierarchical relations not studied previously in schema induction.

\paragraph{Script Learning:} Early script learning work concentrated on chains of events with a single protagonist \cite{chambers-jurafsky-2008-unsupervised, chambers-jurafsky-2009-unsupervised, jans-etal-2012-skip, rudinger-etal-2015-script, Granroth-Wilding_Clark_2016} and later extended to multiple protagonists \cite{pichotta-mooney-2014-statistical, peng-roth-2016-two, Pichotta_Mooney_2016, modi-2016-event, Weber_Balasubramanian_Chambers_2018, weber-etal-2020-causal, zhang-etal-2020-reasoning}. 
All of these works assume there exists a single line of events that describes all occurrences within a complex event.
This work does not limit itself to generating single-chained schemas. We also consider more complex graphs as schema outputs. In addition, none of these works deal with zero-shot scenarios that do not require training data.

\paragraph{Pre-Trained Generation Models:} Large-scale pre-trained text generation models such as GPT-2 \cite{radford2019language}, GPT-3 \cite{brown2020language}, BART \cite{lewis-etal-2020-bart}, T5 \cite{raffel2019exploring}, i.a. have been used in many NLP tasks. These models are often seen as few-shot learners \cite{brown2020language} and therefore used as inference methods.
However, these text generation models are not explicitly trained to perform inference, but to produce the most likely sequence of words to proceed a certain prompt, similar to language models. 
In our work, we use these large pre-trained LMs as text generators.
The generated documents on a particular topic are leveraged as a corpus for extracting the schema of the given topic. 
We rely on the intuition that the generated text will include salient and stereotypical information that is expected to be mentioned in the context of the topic (e.g., for the topic of ``planning a wedding,'' we assume most documents will include ``order catering'').

%% file: 03_data_generation.tex
\section{Data Generation}
\label{03_data_generation}
\input{figures/generated_text}
The schema induction process begins with generating texts using large LMs as text generation models. These texts are joined to form a knowledge base for the schema, including all of the potential information that the schema may present. One could, of course, create this knowledge base by crawling the web for real news articles or Wikipedia entries related to a certain topic. 

We argue, however, that in addition to the obvious advantages of not having to rely on the availability of data online and not having to crawl the entire web for relevant documents on each topic, the generated data from these large generative models is more efficient in reporting salient events than random events described in the news, i.e., generated texts are more likely to mention important information than real documents do.

Our analysis shows that the generated stories contain a higher percentage of relevant tokens than real news articles that are used for schema induction. To demonstrate this phenomenon, we compare manually collected documents with those that are automatically generated using GPT-3 for the event of Improvised Explosive Device (IED) Attack \cite{li-etal-2021-future}. To identify salient events and arguments concerning IED attacks, we adopt the DARPA KAIROS Phase 1 (v3.0) ontology\footnote{The full ontology definition can be accessed at this link:
\url{https://bit.ly/3mIWJoN}.} --- a fine-grained ontology for schema learning, with 24 entity types, 67 event types, and 85 argument roles.

We calculate the number of relevant event triggers and arguments identified in the text, where a relevant mention is one whose type appears in the ontology. The results shown in Table~\ref{tab:generated_vs_real} demonstrate that the quality of the generated texts in terms of conciseness and appearance of important details is higher than that of real texts. For example, the ratio of relevant events per token is more than twice as high in generated texts as it is in real texts.
Hence we are able to not only generate a schema for every given topic without putting any effort in searching the web, but the information we generate is also better suited for our end task of depicting all of the important aspects of a complex event.

\begin{table}[]
\small
\resizebox{\columnwidth}{!}{%
\begin{tabular}{l|c|c}
\hline
\toprule
 & Generated Text & Real Text \\ \hline
\# events / \# tokens & 12.52\% & 6.31\% \\ \hline
\# arguments / \# tokens & 5.45\% & 3.01\% \\ 
\bottomrule
\end{tabular}
}
\caption{The ratio of relevant events and relevant argument roles identified in generated texts and real texts for the scenario of IED attack.}
\label{tab:generated_vs_real}
\vspace{-1em}
\end{table}

Given a topic for which we want to create a schema, we generate multiple texts that discuss the topic event using the OpenAI GPT-3 API\footnote{\url{https://openai.com/blog/openai-api/}.} with the Davinci-instruct-beta-v3 model and we also experiment with the Hugging Face GPT-2 API\footnote{https://huggingface.co/gpt2}. We use three prompting methods to generate documents of diverse genres as follows: 

\stitle{News Articles:} We begin by generating a headline using the prompt: ``Write a news headline about \textit{topic}.'' The output from this prompt is then used in the following prompt: ``Write a news story titled \textit{headline}.'' The output from the second prompt is added to the pool of generated texts. The process is repeated 30 times. See example in \Cref{fig:news_event}.

\stitle{How-To Articles:} We use the prompt: ``Describe how to \textit{topic}.'' to generate wikiHow-like instruction articles. The process is repeated 30 times. See example in \Cref{fig:not_news_event}.

\stitle{Direct Step-by-Step Schema:} We use the prompt: ``What are the steps involved in \textit{topic}? 1.''\footnote{The ``1.'' in the prompt is for the LM to automatically complete the steps.} to directly generate a schema. We run this process once. See example in \Cref{fig:direct_event}. 

Generating documents of various genres enables our model to induce comprehensive schemas on any given topics. Considering that some events are more likely to be in the news (e.g., elections, pandemic outbreaks) while others are more technical in nature and are hence less newsworthy (such as earning a Ph.D. degree or planning a wedding), we generate diverse texts and then use a ranking model to choose the most relevant documents.

The ranking process includes embedding the texts and the topic with the model proposed in \citet{reimers-gurevych-2019-sentence}, and then calculating the cosine similarity between each text and the topic embeddings.
Only the 30 texts closest to the topic are selected,  together with the output from the direct step-by-step schema. The following section describes the next step in generating a schema of extracting relevant information from the texts.

%% file: figures/generated_text.tex
\begin{figure*}
     \centering
     \small
    \begin{subfigure}[t]{0.42\textwidth}
        \raisebox{-\height}{
        \fbox{%
    \parbox{\columnwidth}{%
\textbf{Prompt}: What are the steps involved in holding a summit meeting?\\
1. 

\textbf{Output}:
\begin{enumerate}
    \item \hl{Plan} the agenda and objectives of the meeting.
    \item \hl{Select a venue} that is appropriate for the meeting's purpose and size.
    \item \hl{Send out invitations} and \hl{ensure} all participants have the appropriate information.
    \item \hl{Prepare} for the meeting by \hl{creating materials} and \hl{briefing} the participants.
    \item \hl{Hold} the meeting and \hl{follow up} on any action items that are generated. 
\end{enumerate}
}}}
        \caption{Example generated text of steps involved in the event of ``holding a summit meeting''.}
        \label{fig:direct_event}
        
    \end{subfigure}
    \hspace{1cm}
    \begin{subfigure}[t]{0.47\textwidth}
        \raisebox{-\height}{
        \fbox{%
    \parbox{\columnwidth}{%
    \textbf{First prompt}: Write a news headline about filing a patent.
    
    \textbf{Output}: New Patent Filing Could Speed Up Drug Development.

    \textbf{Second prompt}: Write a news story titled "New Patent Filing Could Speed Up Drug Development".
    
    \textbf{Output}: The United States Patent and Trademark Office has just \hl{published} a new patent that could speed up the drug development process. The patent, \hl{filed by} a team of researchers from the DPharm Institute, \hl{covers} a new drug screening process that could help \hl{identify} potential new drugs much more quickly. The new process, which is based on a computer algorithm, is said to be up to 10 times faster than traditional methods. It can identify potential new drugs by \hl{screening} through large databases of existing drugs and drug candidates. The DPharm Institute is already in the process of \hl{licensing the patent} to several pharmaceutical companies, and is planning to \hl{commercialize the technology} over the next few years.}}
    }
    \caption{Example generation of news-like text for the event of ``filing a patent''.}
    
    \label{fig:news_event}
    \end{subfigure}
    \vspace{1ex}
    \begin{subfigure}[t]{0.9\textwidth}
        \raisebox{-\height}{
        \fbox{%
    \parbox{\columnwidth}{%
\textbf{Prompt}: Describe how to 3D print an object?

\textbf{Output}: To 3D print an object, you need to first \hl{design the object} in a 3D modeling program. Once the object is designed, you need to \hl{export} it as a 3D printable file format. Next, you need to \hl{choose} a 3D printer and print settings. Finally, you need to \hl{print} the object.
}}}
        \caption{Example generation of How-To article for the event of ``3D printing''.}
        \label{fig:not_news_event}
        \end{subfigure}
    \caption{Examples of generated texts using different prompting methods. The highlighted tokens display relevant events that will be extracted in the information extraction step.}
    \label{fig:all_generated}
\end{figure*}

%% file: 04_information_extraction.tex
\section{Information Extraction}
\label{04_information_extraction}

For each document, we extract event triggers, arguments and relations between the events that are important and relevant to the schema topic. We do not work with a predefined ontology that defines what events and arguments are salient in advance because we allow generating a schema on any topic. Instead, we employ a statistical approach by extracting all the information and later filter it down to include just frequent items. Here are the steps involved in our information extraction pipeline:
    \paragraph{Semantic Role Labeling (SRL):} We use the SOTA SRL system\footnote{\label{footnote:SOTA_SRL_NER}
    \url{https://cogcomp.seas.upenn.edu/page/demo_view/SRLEnglish}
    }
    trained on CoNLL12 \cite{pradhan-etal-2012-conll} and Nombank dataset \cite{meyers-etal-2004-annotating} to extract both verb and nominal event triggers and arguments.
    \paragraph{Named Entity Recognition (NER):} 
    We employ the SOTA NER model \cite{guo-roth-2021-constrained} 
    to extract and map entities (potential arguments of events) into entity types defined in the CoNLL 2002 dataset \cite{tjong-kim-sang-2002-introduction} and the LORELEI project \cite{strassel-tracey-2016-lorelei}.
    \paragraph{Constituency Parsing:} The arguments extracted by SRL can be clauses and long phrasal nouns, hence we employ the AllenNLP\footnote{\url{https://demo.allennlp.org/constituency-parsing}.} constituency parsing model for argument head word extraction. 
    \paragraph{Coreference Resolution:} We use the SOTA model \cite{yu-etal-2022-pairwise} for event and entity coreference resolution to identify within-document coreferential relations. 
    \paragraph{Temporal Relation Extraction:} We first try to use SOTA models \cite{ning-etal-2019-improved, zhou-etal-2021-temporal} to predict the temporal relations\footnote{The possible temporal relations (start-time comparison) are: \textsc{Before}, \textsc{After}, \textsc{Equal} and \textsc{Vague}.} between all possible pairs of extracted events but since the SOTA models accept two sentences containing events as input, the inference time\footnote{The inference time is mostly spent on obtaining the contextual representation of events using large fine-tuned LMs.} for an $n$-event document is $\mathcal{O}(n^2)$, making the schema induction process several hours long. 
    
    \paragraph{One-Pass Model:} We develop a One-Pass model that takes the document as input and uses the contextual representation of events to predict relations between them. 
    A document $D$ is represented as a sequence of tokens $D = [t_1, \cdots, e_1, \cdots, e_2, \cdots, t_n]$ where some of the tokens belong to the set of annotated event triggers, i.e., $\mathcal{E}_D = \{e_1, e_2, \cdots, e_k\}$, whereas the rest are other lexemes. We employ the transformer-based language model Big Bird \cite{zaheer2020bigbird} to encode a whole document and obtain the contextualized representations for all the event mentions. These representations are fed into a multi-layer perceptron in a pairwise fashion and the cross-entropy loss for each pair is calculated and accumulated for a batch of documents. As shown in~\Cref{tab:speed}, the inference time is shortened 63-186 times on average, while the performance of the One-Pass model is comparable to SOTA models.
    \paragraph{Hierarchical Relation Extraction:} The extremely long inference time of SOTA models for predicting hierarchical relations (\textsc{Parent-Child}, \textsc{Child-Parent}, \textsc{Coref}, \textsc{NoRel}) \cite{zhou-etal-2020-temporal, wang-etal-2021-learning-constraints} also impairs the efficiency of our schema induction system. Thus we use the same One-Pass methodology to extract hierarchical relations. We observe that the inference time is greatly shortened, and the One-Pass model achieves comparable results to previous models while taking up less GPU memory (see~\Cref{tab:speed}).
    

\begin{table}[]
\resizebox{\columnwidth}{!}{%
\begin{tabular}{l|l|ccc}
\hline
\toprule
                       &                           & \multicolumn{3}{c}{Metrics} \\ 
Corpus                 & Model                                        & $F_1$ score     & Speed  & GPU Memory  \\ \hline
\multirow{3}{*}{HiEve}  & \citet{zhou-etal-2020-temporal}              & 0.489  & -      & -          \\
                        & \citet{wang-etal-2021-learning-constraints}  & 0.522  & 41.68s  & 4515MiB    \\ 
                        & One-Pass model                                & 0.472  & 0.65s  & 2941MiB    \\ \hline
\multirow{3}{*}{MATRES} & \citet{ning-etal-2019-improved}              & 0.767  & 30.12s      & 4187MiB          \\
                        & \citet{zhou-etal-2021-temporal}              & 0.821  & 89.36s & 9311MiB    \\
                        & One-Pass model                                      & 0.768  & 0.48s  & 2419MiB    \\ 
                       \bottomrule
\end{tabular}
}
\caption{Performance comparison between the One-Pass model and SOTA models for event temporal and hierarchical relation extraction. We report $F_1$ scores on benchmark datasets (HiEve for hierarchical relations, MATRES for temporal relations), speed (average inference time for 100 event pairs), and required GPU memory during inference. The One-Pass models are 63-186 times faster than SOTA models and take up only 26\%-65\% of the GPU memory required by SOTA models. }
\label{tab:speed}
\vspace{-0.5em}
\end{table}

After processing the data using the procedure described above, we get a list of events, their arguments, and relations between the events. We concentrate on events and relations that frequently appear in the generated texts since we assume those are the most important to add to the schema (without any other source of information that could identify what is salient). We describe the process of building a schema in the following section.





%% file: 05_schema_induction.tex
\section{Schema Induction}
\label{05_schema_induction}
To consolidate the information extracted from the previous step, we build a schema as follows:
\input{figures/logical_example}

\stitle{Make a list of events and relations}: To compare similar event mentions in different texts, we compare the event trigger itself (whether they are the same verb or coreferential verbs\footnote{\label{footnote:coref-hier}We only consider coreferential and hierarchical relations if they appear in more than 2 documents.}) and the NER types of its arguments. For example, the trigger ``(take) precautions'' appeared in 5 documents generated for the topic of Pandemic Outbreak. In two documents the subject of the verb phrase ``take precautions'' was ``residents'', in another two it was ``people'' and in the last one, it was ``public''. Nevertheless, the NER type is identical in all cases (PER), and thus we set the frequency of ``(take) precautions'' to 5. Similarly, we calculate the frequency of the temporal and hierarchical relations. 
We only consider relations and events that appeared in more than one document.
\input{figures/temp_subevent_example}

\stitle{Construct timelines}: We construct the longest timelines from the list of temporal relations. This list is a list of tuples $(A, B)$, indicating that event $A$ happened before event $B$. To construct a timeline, we search recursively for the longest chains of the following form $(A, B), (B, C), (A, C)$ and so on. 

\stitle{Fix timelines according to hierarchical relations}: We build a hierarchy of the events using the hierarchical relation list\footref{footnote:coref-hier} and change the timelines so that they will only include events that appear in the same level of hierarchy (see example in \Cref{fig:subevent}).

\stitle{Add logical relations}: The final step is to combine the timelines and hierarchies into a single schema graph using logical relations (AND/OR). When observing two timelines with discrepancies between the order of events, we place a logical AND between them, since we interpret this discrepancy as both events occurring at the same time or there is no significance to the order between them. We use a logical OR to mark events that can occur simultaneously but not necessarily. See \Cref{fig:logical} for example of both logical relations.

The final output is a schema graph that contains all the events, arguments, and temporal, hierarchical and logical relations between the events. It is noteworthy that our proposed schema generation model can be easily used to \emph{extend the scope of existing schemas} by further querying the model on more specific topics. For example, the schema in \Cref{fig:kidnapping} does not cover the consequences of kidnapping, probably because the LM did not attend to this aspect. Hence an analyst can input another topic (e.g., consequences of kidnapping) to further develop the schema. Similarly, analysts can generate schemas for very specific events (e.g., kidnapping in a political setting). Next, we provide an in-depth experimentation for the proposed schema induction framework.


%% file: figures/logical_example.tex
\begin{figure*}

\centering
\scalebox{0.8}{
\smartdiagramset{
sequence item font size=\small,
set color list={gray!40, orange!40, orange!40, gray!40, gray!40},
back arrow disabled=true,
text width=2cm
}
\smartdiagram[flow diagram:horizontal]{%
begin unrest, protesters call demands, protesters clash officers, police disperses crowd, condemn violence%
}
}

\vspace{0.1cm}
\scalebox{0.8}{
\smartdiagramset{
sequence item font size=\small,
set color list={gray!40, orange!40, orange!40, blue!40, gray!40},
back arrow disabled=true,
text width=2cm
}
\smartdiagram[flow diagram:horizontal]{%
begin unrest, protesters clash officers, protesters call demands, police disperses crowd, condemn violence%
}
}

\vspace{0.1cm}
\scalebox{0.8}{
\smartdiagramset{
sequence item font size=\small,
set color list={gray!40, gray!40, gray!40, blue!40},
back arrow disabled=true,
text width=2cm
}
\smartdiagram[flow diagram:horizontal]{%
begin unrest, protesters call demands, protesters clash officers, government urges to exercise restraint%
}
}

\vspace{0.1cm}
\scalebox{0.8}{
\smartdiagramset{
sequence item font size=\small,
uniform color list=gray!40 for 4 items,
back arrow disabled=true,
text width=2cm
}
\smartdiagram[flow diagram:horizontal]{%
begin unrest, protesters clash officers, protesters call demands, government urges to exercise restraint}
}

\centering
\begin{tikzpicture}
\draw [thick,dashed] (0,0) -- (7,0);
\end{tikzpicture}
\vspace{0.1cm}

\centering
\scalebox{0.8}{
\smartdiagramset{
sequence item font size=\small,
uniform color list=gray!40 for 5 items,
back arrow disabled=true,
text width=2cm
}
\smartdiagram[flow diagram:horizontal]{%
begin unrest, protesters clash officers \textcolor{orange}{AND} protesters call demands, government urges to exercise restraint \textcolor{blue}{OR} police disperse crowd, condemn violence%
}
}

\caption{An example of integrating timelines and logical relations in the schema of Civil Unrest. The four upper timelines are the ones extracted from the generated texts and the lower one is their merger into a single timeline with logical relations.}
    \label{fig:logical}
\end{figure*}
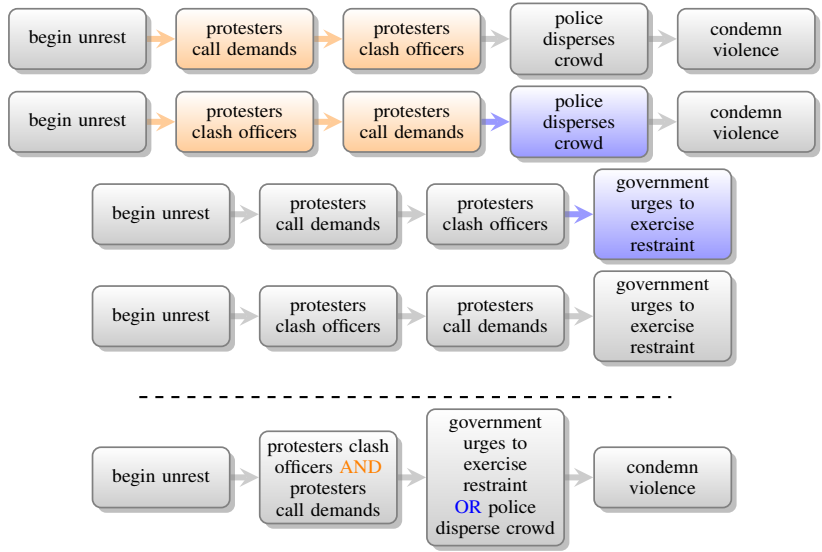

%% file: figures/temp_subevent_example.tex
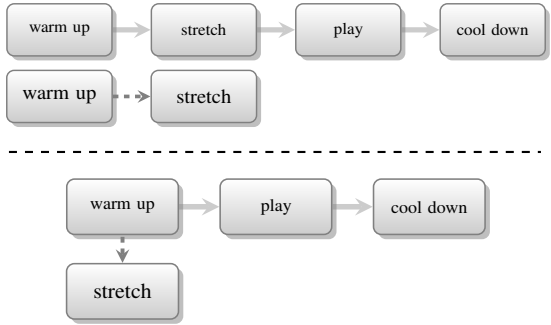
\begin{figure}
    \centering
\begin{center}
\begin{adjustbox}{width = \columnwidth}
\smartdiagramset{
uniform color list=gray!40 for 4 items,
back arrow disabled=true,
additions={
additional item bottom color=gray!40,
additional item shadow=drop shadow,
additional item border color=gray,
additional arrow tip=stealth,
additional arrow line width=2pt,
additional arrow style=dashed,
additional arrow color=gray,
}
}
\smartdiagramadd[flow diagram:horizontal]{%
warm up,stretch,play,cool down%
}{%
below of module1/warm up, below of module2/stretch%
}
\smartdiagramconnect{->}{additional-module1/additional-module2}
\end{adjustbox}

\vspace{0.8cm}
\begin{tikzpicture}
\draw [thick,dashed] (0,0) -- (7,0);
\end{tikzpicture}
\vspace{0.3cm}

\begin{adjustbox}{width = 0.8\columnwidth}
\smartdiagramset{
uniform color list=gray!40 for 3 items,
back arrow disabled=true,
additions={
additional item offset=0.5cm,
additional item bottom color=gray!40,
additional item shadow=drop shadow,
additional item border color=gray,
additional arrow tip=stealth,
additional arrow line width=2pt,
additional arrow style=dashed,
additional arrow color=gray,
}
}
\smartdiagramadd[flow diagram:horizontal]{%
warm up,play,cool down%
}{%
below of module1/stretch%
}
\smartdiagramconnect{<-}{additional-module1/module1}
\end{adjustbox}
\end{center}
\vspace{1cm}
    \caption{Example of the procedure to amend a timeline in the schema of ``Sports Games''. 
    The timeline at the top that includes events from different levels (``warm up'' is the parent of ``stretch'') is fixed below. Gray arrows mark temporal relations, and dashed arrows mark \textsc{Parent-Child}.}
    \label{fig:subevent}
    \vspace{-0.5em}
\end{figure}

%% file: 06_experiments.tex
\section{Experiments}
\label{06_experiments}
\subsection{Data}
We conduct experiments on a dataset for general schema learning released by LDC (LDC2020E25).
The corpus includes 84 types of complex events, such as Cyber Attack, Farming and Recycling. This dataset includes ground-truth schemas created by LDC annotators.
In addition, we also collected human generated schemas for 11 newsworthy scenarios\footnote{The topics are: Bombing Attack, Business Change, Civil Unrest, Disaster and Rescue, Elections, International Conflict, Kidnapping, Mass Shooting, Pandemic Outbreak, Sports Games, and Terrorism Attack.}. The schemas were generated by four human experts who were instructed to write a schema on each topic based on their commonsense knowledge that includes a list of event triggers, event arguments and their NER types\footnote{The annotators are familiar with SRL annotations (e.g., ARG0, ARG1, etc.) and NER types (e.g., PER, ORG, etc.). See additional details in App.~\ref{sec:instruct}}, and relations\footnote{No restrictions were placed for the annotators. For example, in one case, an annotator mentioned causal relations that are not covered in our framework.}.

\subsection{Evaluation Protocols}

We follow \citet{li-etal-2021-future} to use instance coverage and last event prediction to evaluate our method on the LDC dataset. For the Schema-11 dataset, we ask human testers to assess the completeness and soundness of both human- and automatically-generated schemas.

\stitle{Coverage and Prediction}
A common evaluation method in schema induction and script prediction is to calculate the recall of events and relations predicted by the model, assuming the human annotations are gold labels (coverage), and to calculate the accuracy in predicting the final outcome of a scenario (prediction). 
For instance, the accuracy of predicting the last event type of the LDC schemas is reported in \citet{li-etal-2021-future}.
Here we present the results of predicting the last events using event triggers instead of event types since our schemas do not use an ontology of event types. 

\stitle{Feigenbaum Test}
We show human testers two schemas on each topic in the Schema-11 dataset (see example in Appx. \Cref{app:test_details}). One schema is automatically generated by our model, and the other is randomly sampled from the Schema-11 corpus\footnote{In some cases we combine two randomly sampled schemas because the length of the human schemas tend to be shorter than the automatically generated ones.}. 
Then, we ask the testers to determine which events and relations are valid to appear in the schema (soundness), and answer the following questions: which schema is more complete in the sense of including all the events needed to describe the topic, and which schema, in their opinion, was generated by a human expert (as opposed to a machine).


\begin{table*}[]
\centering
{\small
\begin{tabular}{l|cc|cc|c}
\hline
\toprule
& \multicolumn{2}{c|}{GPT2} & \multicolumn{2}{c|}{\textbf{GPT3}} & \citet{li-etal-2021-future} \\ 
& Coverage & Coverage (Syn) & Coverage & Coverage (Syn) & Coverage \\ \hline
Event Match & 14.88 & 29.55 & 23.73 & 37.84 & 54.84 \\ \hline
Temporal Relations & 10.80 & 33.31 & 31.07 & 49.99 & \multirow{3}{*}{-} \\ 
Hierarchical Relations & 33.33 & 33.33 & 11.11 & 13.88  \\ 
Logical Relations & 4.16 & 24.99 & 43.76 & 49.81 \\
\bottomrule
\end{tabular}
}
\caption{Coverage results for the LDC dataset. The first row presents the percentage of events that appeared in both the LDC schemas and the automatically generated schemas (out of events in LDC schemas), and the three bottom rows present the same metric for relations of different types.
}
\label{tab:results_coverage}
\end{table*}

\begin{table}[]
\centering
\small
\begin{tabular}{l|c}
\hline
\toprule
Model & Accuracy \\ \hline
Event Language Model & 49.7 \\ \hline
Sequential Pattern Mining & 47.8 \\ \hline
Human Schema &  20.5\\ \hline
Event Graph Model &  52.0 \\ \hline \hline
Zero-Shot Schema GPT2 & 25.0 \\ \hline
Zero-Shot Schema Synonym GPT2 & 45.2 \\ \hline
Zero-Shot Schema GPT3 & 35.7 \\ \hline
Zero-Shot Schema Synonym GPT3 & \textbf{63.1} \\ 
\bottomrule
\end{tabular}
\caption{Experimental results for last event prediction in the LDC dataset. 
The top 4 results are from \cite{li-etal-2021-future}, and the metric is HITS@1 where the events are typed based on a predefined ontology. }
\label{tab:results_prediction}
\vspace{-0.5em}
\end{table}

\subsection{Results}
\paragraph{Coverage}
We calculate the intersection between events in the generated schemas and the gold schemas in two ways: (a) the matching of event triggers, and (b) the matching of event triggers and synonyms of the events in the gold schemas (synonym coverage)\footnote{Implemented using the NLTK WordNet Python package.}. We believe that synonym coverage is a better evaluation metric to avoid errors such as considering different verbs describing the same action as different (e.g., ``buy'' and ``acquire'') than using a predefined ontology of event types such as the one used in \citet{li-etal-2021-future}. The reason is twofold: firstly, any predefined ontology is limited to certain scenarios and it may impair the variety of events extracted;
and secondly the typing mechanism may also inflict errors to the schema. In the calculation of coverage of relations we only take into account relations $(a,b)$ where both events, $a$ and $b$, appear in the generated schema.

From the results in Table~\ref{tab:results_coverage}, we observe that despite the difficulty of exact matching, our model with GPT-3 covers 23.73\% of the gold events, showing that generated texts are useful. If we use synonym coverage as our metric, we achieve a promising coverage of 37.84\% while the SOTA \emph{supervised} event graph model \cite{li-etal-2021-future} covers 54.84\% using limited event types. In addition, we calculated an average number of 26.19 additional events that appeared in the generated schemas and not in the LDC schema, pointing to the potential of using generated documents for expending existing schemas.
With the high quality event representations obtained from the One-Pass model and the proposed logical relation induction algorithm, our method can successfully cover a high percentage of multiple types of relations. 

\paragraph{Prediction} In the prediction task, our schemas are able reach SOTA performance and predict the final outcome in 63.1\% of the cases for the LDC schemas (see \Cref{tab:results_prediction}). This result is extremely impressive when it is compared with \citet{li-etal-2021-future} since they predict event types instead of verbs, which is a much easier task due to the fact that the set of possible answers is limited. 

\paragraph{Schema-11} In the soundness experiments, where the testers are asked to decide which events and relations are valid to appear in the schema, it turns out that human-schemas contain 7.14\% invalid events and 15.4\% invalid relations on average. For the automatically-generated schemas, 6.06\% of the events and 22.9\% of the relations are considered to be invalid on average, meaning that the average percentage of legitimate events is even higher in machine-generated schemas. 
For the completeness results presented in \Cref{tab:completeness}, in 4 cases the testers agreed that the automatically generated schemas are more complete; in 3 cases they claimed that the human schemas are more complete; and a tie in the remaining 4 cases. Hence our automatically generated schemas are of comparable quality to human generated ones in the sense of completeness.

Finally, in the Feigenbaum test, where testers are asked to decide whether a schema is generated by a human or a machine, eight out of eleven times they correctly identify the human-generated schema, one incorrectly, and two ties. Some of the testers who succeeded in their guesses mentioned that it was easy to determine which schema was automatically generated since it tends to be longer and comprehensive. 
The full results from the Feigenbaum test are shown in Appx. \Cref{app:test_results}.


\begin{table}[]
\resizebox{\columnwidth}{!}{
\begin{tabular}{|l|c|c|c|c|c|c|c|c|c|c|c|}
\hline
 & S1 & \textbf{S2} & \textbf{S3} & \textbf{S4} & S5 & S6 & S7 & S8 & \textbf{S9} & S10 & S11 \\ \hline
Human & 4 & 0 & 1 & 1 & 1 & 2 & 1 & 1 & 0 & 3 & 1 \\ \hline
Automatic & 2 & 3 & 4 & 2 & 1 & 1 & 1 & 1 & 4 & 0 & 1 \\ \hline
\end{tabular}
}
\caption{Distribution of votes for which is the more complete schema for Schema-11 dataset.}
\label{tab:completeness}
\vspace{-0.5em}
\end{table}

\paragraph{Wizard of Oz Experiment}
There seems to be a discrepancy between the low event coverage results and the quality of generated texts that were presented in Section~\Cref{03_data_generation}.
We, therefore, conducted another experiment to identify if the problem stems from the quality of the generated documents. In this experiment, one of the authors sampled 10 complex event names from the LDC dataset and generated, using GPT-3 text davinci-002 model, 3 texts for each scenario using the prompting methods presented in Section~\Cref{03_data_generation}. Then, the author manually extracted all relevant events and relations from each document and built a schema based solely on those events and relations.

This experiment, in which the author pretends to be the IE and schema generator models, aims to demonstrate that if we had perfect IE and schema induction systems, then the texts generated by GPT-3 would be sufficient and even superior to other corpora collected manually. The macro-average coverage of events in this experiment is 68\% and the micro-average is 74\%. Furthermore, GPT-3 texts generated schemas that included, on average, 6.5 additional events not mentioned in LDC schemas but relevant to the scenario at hand. As a result, we conclude that the generated texts from GPT3 contain much of the necessary information to generate schemas in a variety of topics, and can even be used to enrich existing schemas generated by other models or humans. Two example scenarios and more details appear in Appx.~\Cref{sec:app_oz}.

%% file: 09_conclusions.tex
\section{Conclusion}
We propose a method to generate schemas given the sole input of a topic. We use GPT-3 to generate texts of diverse genres and a pipeline of information extraction tools to obtain relevant information before inducing logical relations and integrating the events and relations into a schema graph. To improve the efficiency of the pipeline, we implement One-Pass models for identifying temporal and hierarchical relations that achieve comparable performances with SOTA models but require far less inference time and memory space. To evaluate our framework, we conduct experiments on a benchmark LDC dataset to show that our schemas cover a decent amount of pertinent information and display comparable ability for event prediction with supervised approaches. 
We observe a high percentage of valid events and relations generated for the Schema-11 dataset and the testers endorsed the completeness of our machine-generated schemas.


\label{09_Conclusions}

%% file: 08_ack.tex
\section{Acknowledgments}
\label{08_ack}

The authors would like to thank the anonymous
ACL ARR reviewers for their insightful feedback on our work.
This work was supported by Contract FA8750-19-2-1004 with the US Defense Advanced Research Projects Agency (DARPA). Approved for Public Release, Distribution Unlimited. 
This research is also based upon work supported in part by the Oﬃce of the Director of National Intelligence (ODNI), Intelligence Advanced Research Projects Activity (IARPA), via IARPA Contract No. 2019-19051600006 under the BETTER Program. 
The views and conclusions contained herein are those of the authors and should not be interpreted as necessarily representing the oﬃcial policies, either expressed or implied, of ODNI, IARPA, the Department of Defense, or the U.S. Government. The U.S. Government is authorized to reproduce and distribute reprints for governmental purposes notwithstanding any copyright annotation therein.

%% file: 11_limitations.tex
\section{Limitations}
The paper presents a method for building an event schema without manually collecting documents from sources such as news articles or Wikipedia. In order to generate diverse and informative documents on any topic, we rely on large pre-trained language models. Our model, which uses GPT-3, generates schemas that are comparable to those generated by manually searching the web for documents, however, when we use inferior LMs such as GPT-2, we see a decline in performance (see~\Cref{tab:results_coverage} and~\Cref{tab:results_prediction}).

Our assumption is that the quality of the generated schema depends on the quality of the LM and the level of coverage of the selected topic in the LM training data. If, for instance, we were to ask our model to generate a schema for a unique topic such as "conducting an archaeological dig in an unexplored territory" we doubt that the results would be as useful to an archaeologist as if they were looking for information themselves due to the low coverage of this topic in the corpus the model was trained on. Despite our model's reliance on pre-trained LMs, we believe the generated schemas can always serve as a good basis for further development.

%% file: 10_ethics.tex
\section{Ethical Consideration}
\label{10_ethics}
The proposed schema induction method does not present any direct societal implications. As is observed in \citet{abid2021large}, the text generated by GPT-3 might include undesired social bias. Extracting events and relations from text with such social bias might potentially propagate the bias to the induced schemas. Besides, there are risks of malicious or unintended harmful uses of the generated schemas, for instance, the system might be used to inquire about making a bomb or contriving a terrorist attacks. Yet we believe that the proposed method can benefit various downstream NLP/NLU tasks like event prediction, task-oriented dialogue agents \cite{andreas-etal-2020-task} and risk detection \cite{pohl2012automatic}.

%% file: appendix_turing_experiment.tex
\section{Feigenbaum Test Details}
\label{app:test_details}
\begin{figure*}
    \centering
    \includegraphics[width=\textwidth]{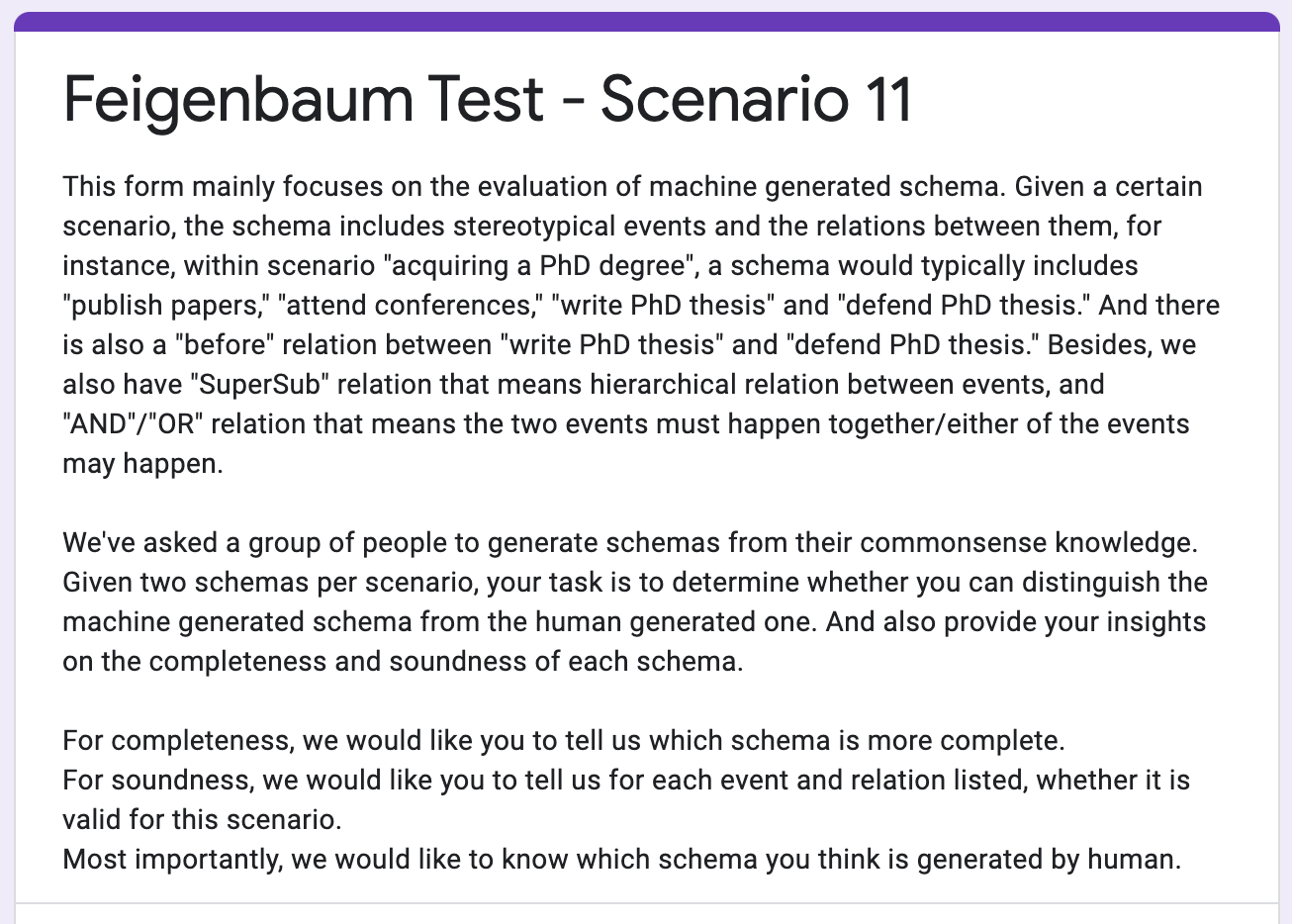}
    \caption{Instructions for the Feigenbaum test.}
    \label{fig:test_explanation}
\end{figure*}

\begin{figure*}
    \centering
    \includegraphics[width=\textwidth]{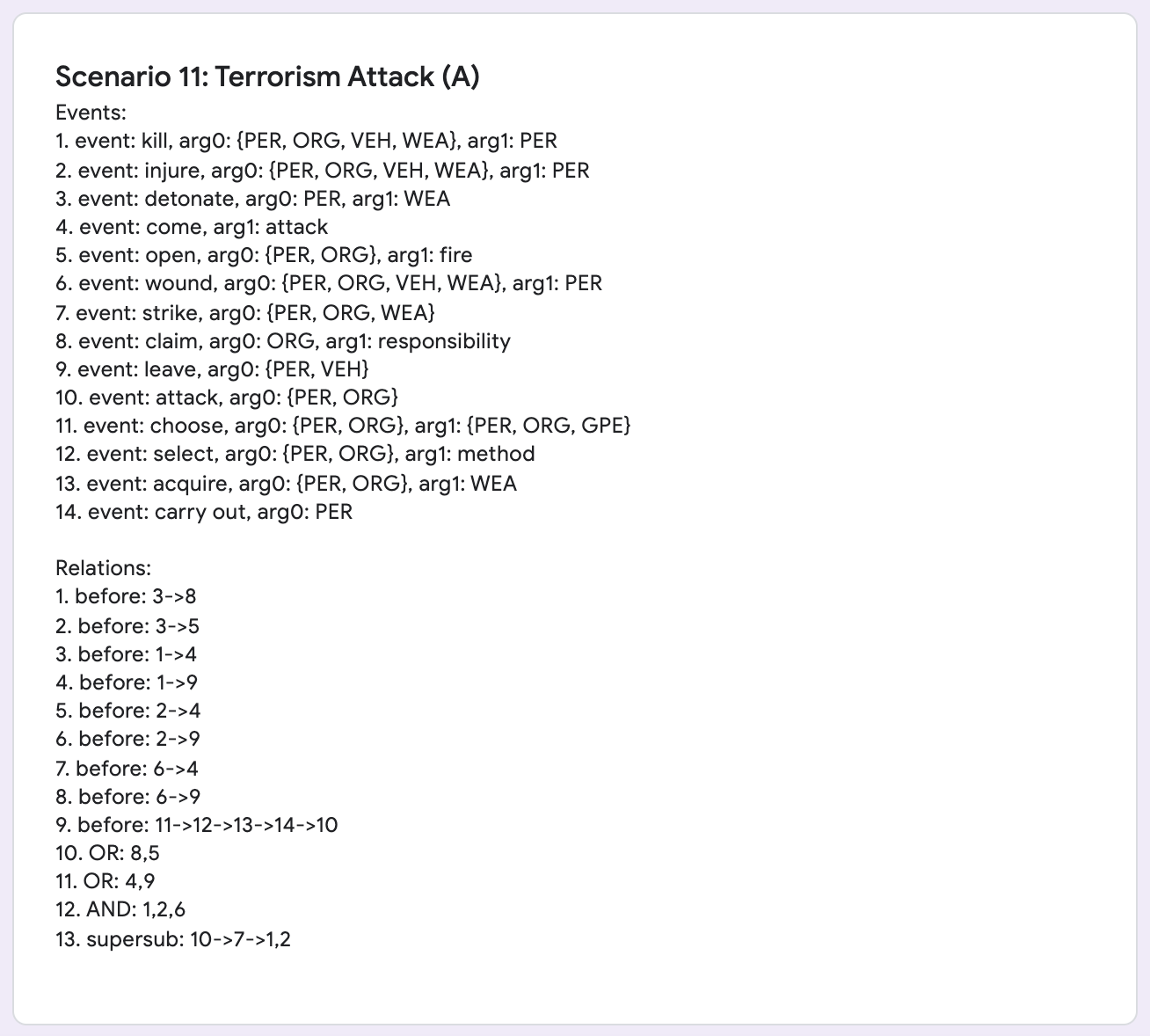}
    \caption{An example schema in the topic of Terrorism Attack. This schema was generated automatically (information that was unknown to the annotators).}
    \label{fig:test_schemaA}
\end{figure*}

\begin{figure*}
    \centering
    \includegraphics[width=\textwidth]{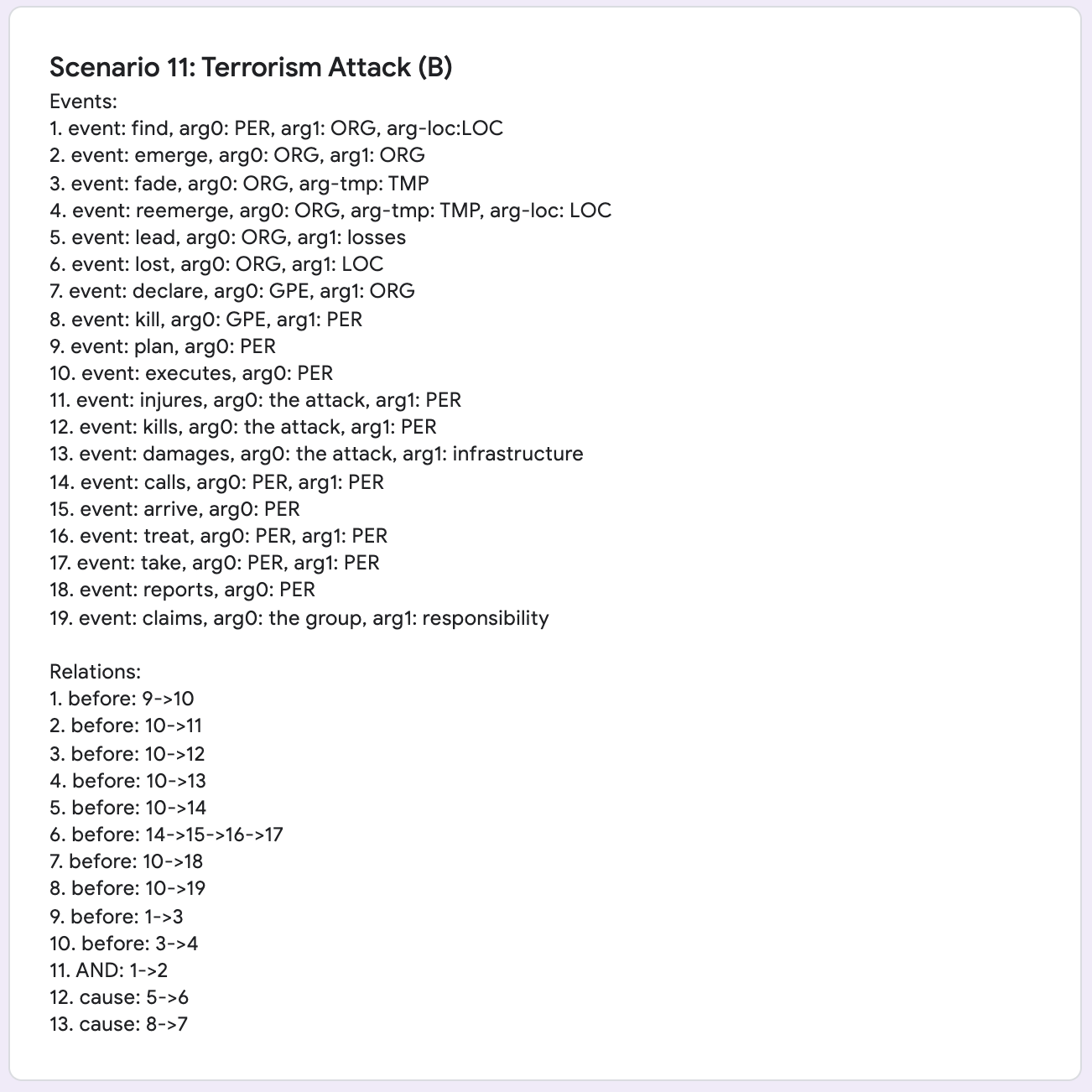}
    \caption{An example schema in the topic of Terrorism Attack. This schema was generated by a human (information that was unknown to the annotators). }
    \label{fig:test_schemaB}
\end{figure*}

\begin{figure*}
    \centering
    \includegraphics[width=\textwidth]{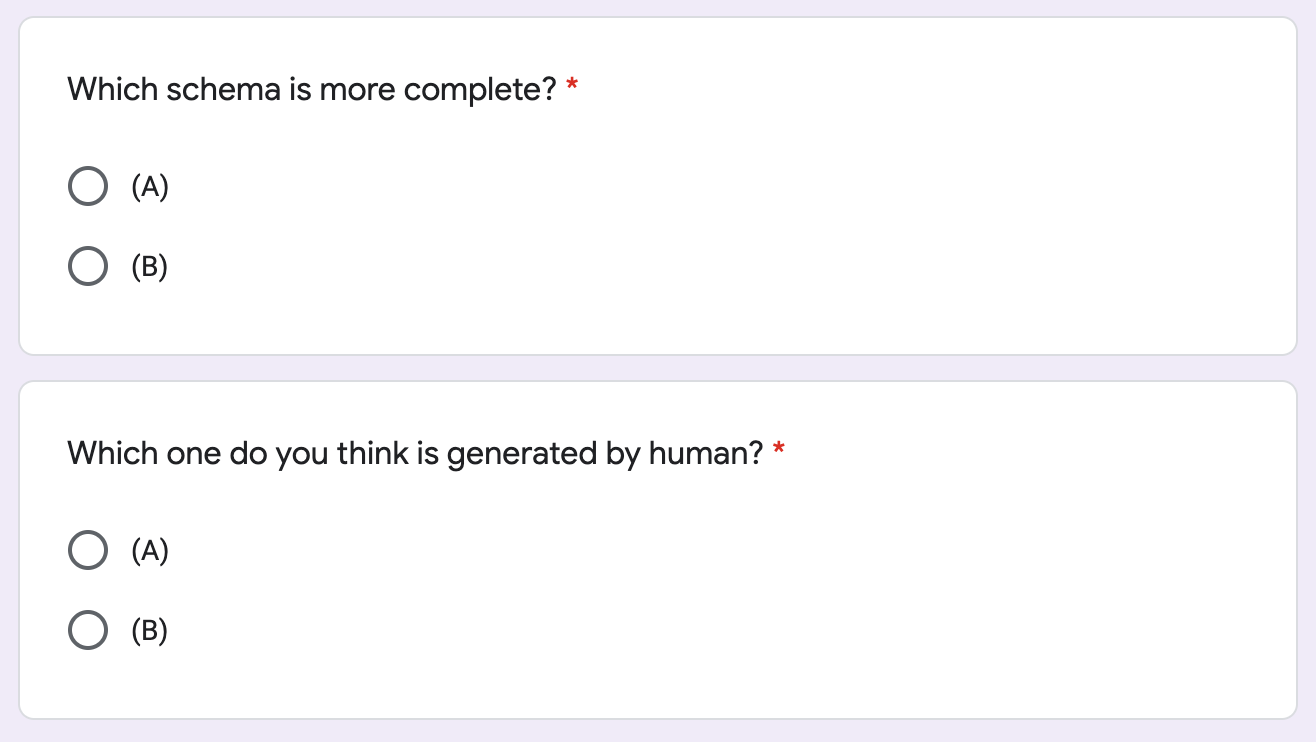}
    \caption{Questions that were asked about the completeness of the schemas and the generator of the schema.}
    \label{fig:test_q}
\end{figure*}

\begin{figure*}
    \centering
    \includegraphics[width=\textwidth]{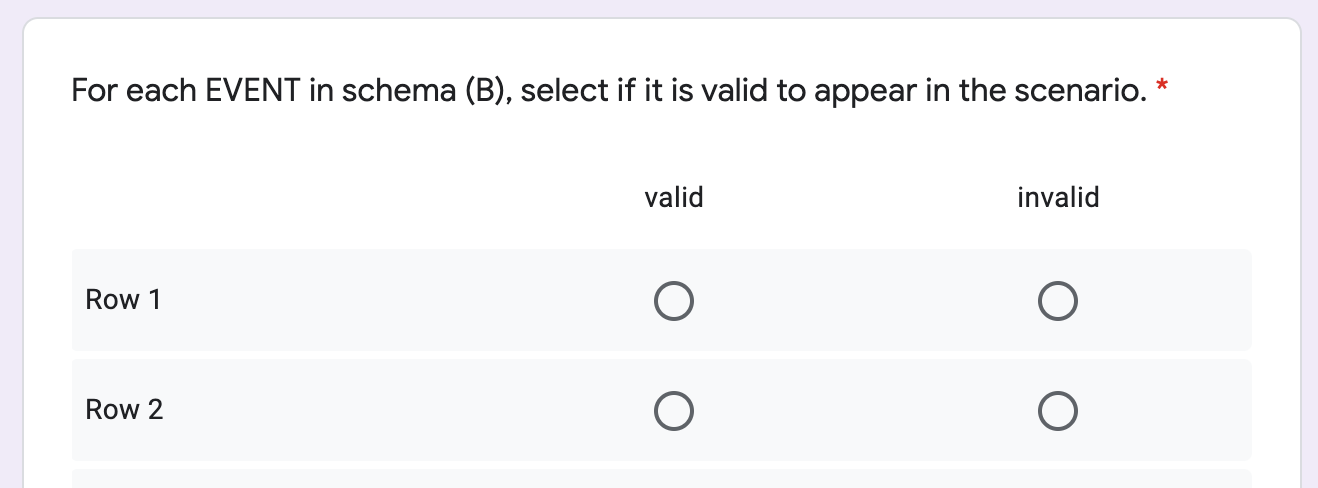}
    \caption{Questions about the validity of the events appearing in one of the schemas. This question was asked on both schemas and on the relations appearing in the schemas too.}
    \label{fig:test_q1}
\end{figure*}

The experiment took place online through filling a Google Form and involved 11 volunteer annotators. Each annotator got 3-4 scenarios to annotate. The instructions for the survey appear in Figure~\ref{fig:test_explanation}. An example scenario and the questions of the survey are presented in~\Cref{fig:test_schemaA}, \Cref{fig:test_schemaB}, \Cref{fig:test_q}, and \Cref{fig:test_q1}.

%% file: App1_turing_results.tex
\section{Feigenbaum Test Results}
\label{app:test_results}
In this section we present all the results from the experiments on the dataset Schema-11. \Cref{tab:completeness_res_schema_11} shows the distribution of answers for the question ``which schema is more complete?'' (same as depicted in \Cref{tab:completeness}), \Cref{tab:turing_res} presents the distribution of answers for the question "which schema was generated by a human?" together with the correct answer written in the bottom row, and \Cref{tab:invalid_res} presents the percentage of invalid events and relations determined by the majority vote of the annotators in the automatic schema and the human schema.

\begin{table*}[]
\resizebox{\textwidth}{!}{%
\begin{tabular}{|l|c|c|c|c|c|c|c|c|c|c|c|}
\hline
 & \multicolumn{1}{l|}{S1} & \multicolumn{1}{l|}{S2} & \multicolumn{1}{l|}{S3} & \multicolumn{1}{l|}{S4} & \multicolumn{1}{l|}{S5} & \multicolumn{1}{l|}{S6} & \multicolumn{1}{l|}{S7} & \multicolumn{1}{l|}{S8} & \multicolumn{1}{l|}{S9} & \multicolumn{1}{l|}{S10} & \multicolumn{1}{l|}{S11} \\ \hline
Human & \hl{4} & 0 & 1 & 1 & 1 & \hl{2} & 1 & 1 & 0 & \hl{3} & 1 \\ \hline
Automatic & 2 & \hl{3} & \hl{4} & \hl{2} & 1 & 1 & 1 & 1 & \hl{4} & 0 & 1 \\ \hline
\end{tabular}
}
\caption{Completeness results. The table presents the number of votes that were recorded for which schema is more complete - the human generated schema or the automatically generated schema. The majority vote is highlighted in yellow.}
\label{tab:completeness_res_schema_11}
\end{table*}

\begin{table*}[]
\resizebox{\textwidth}{!}{%
\begin{tabular}{|l|c|c|c|c|c|c|c|c|c|c|c|}
\hline
 & \multicolumn{1}{l|}{S1} & \multicolumn{1}{l|}{S2} & \multicolumn{1}{l|}{S3} & \multicolumn{1}{l|}{S4} & \multicolumn{1}{l|}{S5} & \multicolumn{1}{l|}{S6} & \multicolumn{1}{l|}{S7} & \multicolumn{1}{l|}{S8} & \multicolumn{1}{l|}{S9} & \multicolumn{1}{l|}{S10} & \multicolumn{1}{l|}{S11} \\ \hline
A & 1 & 1 & 3 & 0 & 0 & \hlc[green]{2} & 0 & \hlc[green]{2} & 2 & 1 & 1 \\ \hline
B & \hlc[green]{5} & \hlc[green]{2} & \hlc[red]{2} & \hlc[green]{3} & \hlc[green]{2} & 1 & \hlc[green]{2} & 0 & 2 & \hlc[green]{2} & 1 \\ \hline
Correct Answer & B & B & B & B & B & A & B & A & A & B & B \\ \hline
\end{tabular}
}
\caption{Feigenbaum test results. The annotators guesses which schema (A or B) was generated by humans. The number of votes for each option appear along with the correct answer in the bottom row. The correct majority guesses are marked with green and incorrect with red.}
\label{tab:turing_res}
\end{table*}

\begin{table*}[]
\resizebox{\textwidth}{!}{%
\begin{tabular}{|l|c|c|c|c|c|c|c|c|c|c|c|}
\hline
 & \multicolumn{1}{c|}{S1} & \multicolumn{1}{c|}{S2} & \multicolumn{1}{c|}{S3} & \multicolumn{1}{c|}{S4} & \multicolumn{1}{c|}{S5} & \multicolumn{1}{c|}{S6} & \multicolumn{1}{c|}{S7} & \multicolumn{1}{c|}{S8} & \multicolumn{1}{c|}{S9} & \multicolumn{1}{c|}{S10} & \multicolumn{1}{c|}{S11} \\ \hline
Invalid Events (Auto.) & 0 & 0 & 0 & 0 & 0 & 8.33 & 0 & 7.69 & 0 & 14.28 & 0 \\ \hline
Invalid Relations (Auto.) & 46.15 & 16.66 & 25 & 25 & 0 & 23.52 & 0.4 & 11.76 & 12.5 & 22.22 & 46.15 \\ \hline
Invalid Events (Human) & 0 & 0 & 14.28 & 14.28 & 0 & 0 & 0 & 0 & 0 & 0 & 0 \\ \hline
Invalid Relations (Human) & 7.69 & 50 & 15.38 & 15.38 & 0 & 6.25 & 0 & 11.11 & 0 & 10 & 7.69 \\ \hline
\end{tabular}
}
\caption{Invalidity results. The table presents the percentage of invalid events and relations determined by the human annotators for each schema and scenario.}
\label{tab:invalid_res}
\end{table*}

%% file: app2_instructions.tex
\section{Details on Human Schema Curation}
\label{sec:instruct}
Here are the instructions that were given to the annotators that generated the human schemas for the Schema-11 dataset. All the annotators are graduate students that previously were involved in research projects that include schema induction, SRL, NER or other relevant tasks:

We are developing a system that generates schemas automatically given a topic. We want to compare our automatically-generated schema to schemas derived by people using their commonsense (without relying on texts).
To do this, we need expert human annotators and would appreciate your assistance. 

A schema is defined as a list of events with their argument types, and the relationships between the events.
For example, here is a schema I wrote that describes the event of "armed robbery": 

List of events and arguments:
\begin{itemize}
    \item intend: arg0 - perpetrator [PER], arg1 - commit a felony
    \item acquire: arg0 - perpetrator [PER], arg1 - weapon [WEA]
    \item arrive: arg0 - perpetrator [PER], arg-loc - crime scene [LOC]
    \item assault: arg0 - perpetrator [PER], arg1 - [PER]
    \item threaten: arg0 - perpetrator [PER], arg1 - [PER]
    \item get: arg0 - perpetrator [PER], arg1 - money or goods
    \item injure: arg0 - perpetrator [PER], arg1 - [PER]
    \item kill: arg0 - perpetrator [PER], arg1 - [PER]
    \item flee: arg0 - perpetrator [PER], arg-loc - crime scene [LOC]
    \item call: arg0 - [PER], arg1 - police [ORG]
    \item chase: arg0 - police [ORG], arg1 - perpertrator[PER]
    \item catch: arg0 - police [ORG], arg1 - perpertrator[PER]
    \item manage to escape: arg0 - perpertrator[PER]
\end{itemize}

Temporal and logical relations (in the form of a timeline):
\begin{itemize}
    \item a perpetrator (PER) \textbf{intent} to commit a felony -> 
    \item the perpetrator (PER) \textbf{acquires} weapon (WEA) -> 
    \item the perpetrator (PER) \textbf{arrives} at the scene (LOC) -> 
    \item perpetrator (PER) \textbf{assault} victim (PER) with weapon (WEA) at the scene (LOC) \textbf{OR} perpetrator (PER) \textbf{threatens} a person (PER) with the weapon (WEA) at the scene (LOC) ->
    \item perpetrator (PER) \textbf{gets} money or goods from the person (PER) \textbf{OR} victim \textbf{injured OR} victim \textbf{killed} -> 
    \item perpetrator \textbf{flees} the scene of the crime (LOC) \textbf{AND} someone (PER) \textbf{calls} the police (ORG) -> 
    \item the police (ORG) are \textbf{chasing} the criminal (PER) -> 
    \item the police (ORG) \textbf{catches} the perpetrator (PER) \textbf{XOR} the criminal (PER) \textbf{manages to escape}.
\end{itemize}

The complex events we are interested in are the following:
(1) Disease Outbreak
(2) IED Bombing
(3) Civil Unrest
(4) International Invasion
(5) Disaster and Rescue
(6) Terrorism Attacks
(7) Election
(8) Kidnapping
(9) Business Change
(10) Mass Shooting.

%% file: wizard_of_oz.tex
\section{Wizard of Oz Experiment}
\label{sec:app_oz}
Two examples of the Wizard of Oz experiment for the events of invasion and remodel a facility. The examples show the three generated texts using the three prompting methods, the schema created by the author, and the LDC schema together with a matching between the events in the two schemas. The entire list of sampled topics is: print a 3D object, cyber attack, make a website, religious conversion, invasion, shipping and mailing, remodel a facility, obtain and provide medical treatment, build a drone, and nonviolent protest march.


\subsection{Example 1: Invasion}
\paragraph{Prompts and Outputs:}
\begin{enumerate}
    \item Prompt: describe how to perform an invasion.\\
    Output: 1. Choose a target: Identify a target population that you want to invade. This target population can be a group of people, a city, or a country. 
2. Prepare your troops: Assemble your troops and supplies in preparation for the invasion. 
3. Launch your attack: Attack the target population using whatever means necessary to take control. 
4. Consolidate your power: Once you have taken control of the target population, consolidate your power by establishing your authority and eliminating any resistance.
\item Prompt: what are the steps involved in an invasion?\\
Output: 1. The first step in an invasion is to gather intelligence about the enemy. This includes learning about their strengths, weaknesses, and their territory. 
2. The next step is to plan the invasion. This includes deciding what resources will be needed and how they will be used. 
3. The third step is to launch the invasion. This is when the actual fighting begins. 
4. The fourth step is to consolidate the gains. This is when the invading force secures their position and prepares for the next phase of the operation.
\item Prompt: write a news headline about invasion.\\
Output: U.S. troops invade Canada in surprise attack.\\
Prompt: write a news article titled: “U.S. troops invade Canada in surprise attack.”\\
Output: In a stunning move, the United States has invaded Canada in a surprise attack. The troops came across the border early this morning, catching the Canadians by surprise. The Canadians are currently fighting back, but it is unknown how long they will be able to hold out against the might of the United States military. This is a developing story, and more details will be released as they become available.
\end{enumerate}
\paragraph{Schema generated based on the three outputs}
\begin{enumerate}
    \item Choose target.
    \item Gather intelligence on target: learn strengths and weaknesses and territory.
    \item Plan invasion: decide on resources and how the resources will be used.
    \item Prepare troops: Assemble troops and supplies.
    \item Launch attack: come across the border.
    \item Fight or use whatever means possible.
    \item Take control.
    \item Consolidate power: establish authority and eliminate resistance.
    \item Prepare for the next phase of operation.
\end{enumerate}

\paragraph{LDC Schema}
\begin{enumerate}
    \item Step 1: PlanInvasion: Plan[3] for the invasion.
    \begin{enumerate}
        \item Step 1.1: IdentifyTargets[1]: Leadership builds a list of entry points for the invasion and selects one or more as the target(s) of the initial invasion.
        \item Step 1.2: DevelopPlan[3]: A team develop a master invasion plan.
        \item Step 1.3: OrderDissemination [not matched]: Planners send out detailed instructions to military forces about the invasion plan.
    \end{enumerate}
    \item Step 2: DeployTroops[4]: Amass and deploy troops for invasion.
    \item Step 3: Invasion[5]: Military force launch the invasion and enter the target territory.
    \item Step 4: DisperseTroops[6]: Dispersion of invading troops across territory.
\end{enumerate}
\paragraph{Results:} The GPT3 schema covers 6 out of 7 events that are mentioned in the LDC schema. The matching between LDC events and schema events are marked in [] next to the event on the LDC schema. GPT3 schema has 7 additional events: gather intelligence, learn strengths, decide on resources, assemble troops and supplies, come across border, take control, consolidate power, establish authority, eliminate resistance, prepare for next phase.

\subsection{Example 2: Remodel A Facility}
\paragraph{Prompts and Outputs:}
\begin{enumerate}
    \item Prompt: describe how to remodel a facility.\\
    Output: There is no one-size-fits-all answer to this question, as the best way to remodel a facility depends on the specific needs and goals of the organization. However, some tips on how to remodel a facility include:
1. Define the scope of the project.
2. Create a budget and timeline.
3. Hire an experienced architect or designer.
4. Work with a reputable contractor.
5. Use high-quality materials.
6. Follow building codes and regulations.
7. Get input from stakeholders.
8. Test the new space before moving in.
    \item Prompt: what are the steps involved in remodeling a facility?\\
    Output: 1. The first step in remodeling a facility is to develop a plan. This plan should include the goals of the remodel, the budget, the timeline, and the scope of work. 
2. The next step is to obtain the necessary permits. Depending on the scope of work, this may require a building permit, a zoning permit, and/or a permit from the local health department. 
3. Once the permits are obtained, the next step is to select a contractor. The contractor should be licensed and insured, and should have experience with the type of work that needs to be done. 
4. The contractor will then begin the work, which may include demolition, construction, painting, and/or installation of new fixtures and appliances. 
5. Once the work is completed, the facility will need to be inspected to ensure that it meets all applicable codes and regulations. 
6. Finally, the facility can be reopened to the public.
    \item Prompt: write a news headline about remodeling a facility.\\
    Output: Facility Remodeling Underway\\
    Prompt: write a news article titled: "Facility Remodeling Underway"\\
    Output: In an effort to better serve the needs of the community, the local library is undergoing a remodel. The project is expected to take several months to complete, but the end result will be a more user-friendly space that meets the needs of today’s patrons. Some of the changes include a new layout for the adult fiction section, additional study rooms, and updated furniture throughout the facility. The children’s area will also be expanded to create a more inviting and interactive space. The library staff is excited about the changes and can’t wait to see the finished product. They are confident that the updates will make the library a more popular destination for both residents and visitors.
\end{enumerate}
\paragraph{Schema generated based on the three outputs}
\begin{enumerate}
    \item Develop a plan: define scope and goals and create budget and timeline.
    \item Hire an architect or designer and obtain necessary permits.
    \item Select a contractor (preconditions: contractor licensed, insured, have experience).
    \item Contractor works: use high quality materials AND follow building code and regulations and demolition or construction or painting or installation.
    \item Get input from stakeholders.
    \item Inspect facility.
    \item Test space.
    \item Reopen facility.
    \item Facility is user-friendly and meets needs of patrons.
\end{enumerate}

\paragraph{LDC Schema}
\begin{enumerate}
    \item Step 1: Acquisition [not matched]: Acquire facility.
    \item Step 2: Planning[1]: Research and plan expected usage, budget, changes, legal issues, dependencies.
    \item Step 3: Labor[2,3]: Arrange for skills, or laborers, or both.
    \begin{enumerate}
        \item Step 3.1: AcquireSkills [not matched]: Acquire skills or knowledge required for remodeling.
        \item Step 3.2: HireLaborers[2,3]: Hire skilled person or organization to perform remodeling work.
    \end{enumerate}
    \item Step 4: AcquireMaterials[4.1]: Acquire materials and tools.
    \item Step 5: Remodel[4]: Facility is remodeled.
    \begin{enumerate}
        \item Step 5.1: Demolition[4.2]: Deconstruction or demolition of portions of building and/or equipment installations.
        \item Step 5.2: DebrisRemoval [not matched]: Hauling away/dumping of debris.
        \item Step 5.3: Modification[4.2]: Modification, addition, or installation of building or systems/equipment in building.
    \end{enumerate}
    \item Step 6: Inspection[6,7]: Inspect and/or test new portions of facility and/or new systems of facility for functionality and compliance with laws and regulations.
\end{enumerate}

\paragraph{Results:} The GPT3 schema covers 8 out of 11 events that are mentioned in the LDC schema. The matching between LDC events and schema events are marked in [] next to the event on the LDC schema. GPT3 schema has 9 additional events: contractor works, follow building code and regulations, preconditions on contractor, painting, installation, construction.